\documentclass{article}

\usepackage[utf8]{inputenc}
\usepackage[margin=1.5in]{geometry}
\usepackage{amsmath}
\usepackage{amsfonts}
\usepackage{amssymb}
\usepackage{mathrsfs}
\usepackage{amsthm}
\usepackage{mathtools}
\usepackage{natbib}
\usepackage[pdftex]{graphicx}
\usepackage{color}
\usepackage{tikz}
\usepackage{comment}
\usepackage{bbm}
\usepackage{multirow}
\usepackage{url}

\definecolor{codegreen}{rgb}{0,0.6,0}
\definecolor{codegray}{rgb}{0.5,0.5,0.5}
\definecolor{codepurple}{rgb}{0.58,0,0.82}
\definecolor{backcolour}{rgb}{0.95,0.95,0.92}

\newtheorem{thm}{Theorem}[subsection]
\newtheorem{defn}[thm]{Definition}

\newcommand{\act}{\sigma}
\newcommand{\be}{\begin{equation}}
\newcommand{\ee}{\end{equation}}
\def \bsp {\begin{split}}
\def \esp {\end{split}}
\def \bea {\begin{eqnarray}}
\def \eea {\end{eqnarray}}

\DeclareMathOperator*{\mini}{minimize}

\title{Deep Fundamental Factor Models}
 
\author{Matthew F. Dixon\thanks{Department of Applied Math, Illinois Institute of Technology, Chicago; matthew.dixon@iit.edu.}
    \and
    Nicholas G. Polson\thanks{ChicagoBooth, University of Chicago, Chicago; ngp@chicagobooth.edu.}
    }

\begin{document}
\maketitle
\begin{abstract}
\noindent Deep fundamental factor models are developed to automatically capture non-linearity and interaction effects in factor modeling. Uncertainty quantification provides interpretability with interval estimation, ranking of factor importances and estimation of interaction effects. With no hidden layers we recover a linear factor model and for one or more hidden layers, uncertainty bands for the sensitivity to each input naturally arise from the network weights. Using 3290 assets in the Russell 1000 index over a period of December 1989 to January 2018, we assess a 49 factor model and generate information ratios that are approximately 1.5x greater than the OLS factor model.  Furthermore, we compare our deep fundamental factor model with a quadratic LASSO model and demonstrate the superior performance and robustness to outliers. The Python source code and the data used for this study are provided\footnote{See \url{https://github.com/mfrdixon/Deep_Fundamental_Factors} for Python notebooks and data.}. 


\end{abstract}



\section{Introduction}

 In this paper, we present a framework for deep fundamental factor (DFF) models. The key aspect is a methodology for interpreting the deep learner which is applicable under mild restrictions on the network architecture. Our method explicitly identifies interaction effects and ranks the importance of the factors. In the case when the network contains no hidden layers, we recover ordinary least squares estimators. For one or more hidden layers, we show how uncertainty bands for the sensitivity of the model to each input arise from the network weights. Moreover, for a certain choice of activation functions, we prove that such a distribution has bounded mean and variance, with variance bounds expressed as a function of the product of weights in the network. Such bounds guarantee finite bootstrapped confidence intervals from the factor sensitivity distribution and provide a valid inference approach for deep networks in contrast to hypothesis tests for neural networks which are limited to shallow networks and use asymptotic estimates.
 

 \paragraph{Deep learning} 
 
Deep learning applies hierarchical layers of hidden variables to construct nonlinear predictors which scale to high dimensional input space. The deep learning paradigm for data analysis is algorithmic rather than probabilistic. Deep learning has been shown to `compress' the input space by projecting the input variables into a lower dimensional space using auto-encoders, as in deep portfolio theory \citep{heaton_deep_2016}. A related approach introduced by \cite{FAN2017292}, referred to as \emph{sufficient forecasting}, provides a set of sufficient predictive indices which are inferred from high-dimensional predictors.  The approach uses projected principal component analysis under a semi-parametric factor model and has a direct correspondence with deep learning.

 \subsection{Why Deep Neural Networks?} 
 
 Artificial neural networks have a long history in financial modeling.  Most recently, the literature has been extended to include deep neural networks (see, for example, \cite{Feng2018,heaton_deep_2016, pelgery2019}). 
 It is well-known that shallow neural networks are furnished with the universal representation theorem, which states that any shallow\footnote{While there is no consensus on the definition of a shallow network, we shall refer to a network with one hidden layer as shallow.} feedforward neural network can represent all continuous functions \citep{Hornik1989}, \emph{provided there are enough hidden units}. It has recently been shown that deep networks can achieve superior performance versus linear additive models, such as linear regression, while avoiding the curse of dimensionality \citep{poggio_deep_2016}. 
\cite{Feng2018} show that deep neural networks provide powerful expressability\footnote{Expressability is a measure of the generality of the class of functions representated by the network. In the context of classification, expressibility is measure by VC dimension.} when combined with regularization, however their use in factor modeling presents some fundamental obstacles, one of which we shall address in this paper, namely interpretability. Neural networks have been presented to the investment management industry as 'black-boxes'. As such they are not viewed as interpretable and their internal behavior can't be reasoned on statistical grounds.  We add to the literature by introducing a method of interpretability which holds for any deep network. One important caveat is that we do not attempt to solve the causation problem in economic modeling.
 
 \paragraph{Non-linearity} While high dimensional data representation is one distinguishing aspect of machine learning over linear regression, it is not alone. Deep learning resolves predictor non-linearities and interaction effects without over-fitting through a bias-variance tradeoff. As such, it provides a highly expressive regression model for complex data which relies on compositions of simple non-linear functions rather than being additive. We develop a deep fundamental factor model with 50 factors for 3290 Russell 1000 indexed stocks over an approximate 30 year period and compare performance and factor interpretability with OLS and LASSO based fundamental factor models.


The rest of the paper is outlined as follows. Section \ref{sect:cff} provides the connection with deep learning and factor models in finance. Section \ref{sect:fund} introduces the terminology and notation for defining neural network based fundamental factor models. Section \ref{sect:interp} introduces our sensitivity based interpretability approach which, in principle, is sufficiently general for fundamental factor models such as the Barra model. We prove that the sensitivity has bounded variance and is therefore useful to characterize uncertainty in factor importance.
Section \ref{sect:results} demonstrates the application of our framework to NN based factor models. Finally, Section \ref{sect:summary} concludes with directions for future research.


 \subsection{Connection with Fundamental Factor Models} \label{sect:cff}
 Barra factor models (see \cite{RePEc:ucb:calbrf:44, Carvalho9}) are appealing because of their simplicity and their economic interpretability, generating tradable portfolios. Under the assumption of homoscedasticity, factor realizations can be estimated in the Barra model by ordinary least squares regression\footnote{The Barra factor model is often presented in the more general form with heteroscedastic error but is not considered here.}. OLS linear regression exhibits poor expressability and relies on the Gaussian errors being orthogonal to the regressors. Generalizing to non-linearities and incorporating interaction effects is a harder task.

 


 
 Asset managers seek novel predictive firm characteristics to explain anomalies which are not captured by classical capital asset pricing and factor models. Recently a number of independent empirical studies, have shown the importance of using a higher number of economically interpretable predictors related to firm characteristics and other common factors \citep{moritz2016, 10.1093/rfs/hhv059, Gu2018, Feng2018}.  \cite{Gu2018} analyze a dataset of more than 30,000  individual  stocks  over a 60 year period from 1957 to 2016, and determine over 900 baseline signals. \cite{moritz2016,pelgery2019,Gu2018} highlight the inadequacies of OLS regression in variable selection over high dimensional datasets - in particular the inability to capture outliers. In contrast, deep neural network can explain more structure in stock returns because of their ability to fit flexible functional forms with many covariates. We build on the work of \cite{Feng2018} that demonstrates the ability of a three-layer deep neural network to effectively predict asset returns from fundamental factors by providing interpretability.  In contrast to \cite{pelgery2019}, we do not attempt to enforce the no-arbitrage constraint but share similiar goals in attempting to rank the importance of the factors. 

\section{Deep Fundamental Factor Models} \label{sect:fund}
\cite{RePEc:ucb:calbrf:44} introduced a cross-sectional fundamental factor model to capture the effects of macroeconomic events on individual securities. The choice of factors are microeconomic characteristics – essentially common factors, such as industry membership, financial structure, or growth orientation \citep{Nielsen2010}. 

The Barra fundamental factor model\footnote{Specifically, the predictive form of the Barra model is conventionally used for risk modeling, however, we shall use adopt this model for a stock selection strategy based on predicted returns.} expresses the linear relationship between $K$ fundamental factors and $N$ asset returns:
\be
\mathbf{r_t} =B_t\mathbf{f}_t + \mathbf{\epsilon}_t,~ t=1,\dots,T,
\ee
where $B_t= [\mathbf{1}~|~{\boldsymbol \beta}_{1}(t)~|\cdots|~{\boldsymbol \beta}_{K}(t)]$ is the $N\times K+1$ matrix of known factor loadings (betas): $\beta_{i,k}(t):=\left(\boldsymbol{\beta}_k\right)_i(t)$ is the exposure of asset $i$ to factor $k$ at time $t$. The factors are asset specific attributes such as market capitalization, industry classification, style classification. $\mathbf{f}_t=[\alpha_t, f_{1,t},\dots, f_{K,t}]$  is the $K+1$ vector of unobserved factor realizations at time $t$, including $\alpha_t$. $\mathbf{r_t}$ is the $N-$vector of asset returns at time $t$. The errors are assumed independent of the factor realizations $\rho(f_{i,t}, \epsilon_{j,t})=0, \forall i,j, t$ with homoschedastic Gaussian error, $\mathbb{E}[\epsilon^2_{j,t}]=\sigma^2.$

\subsection{Deep Factor Models}
Consider a non-linear cross-sectional fundamental factor model of the form
\be \label{eq:non-linear}
\mathbf{r}_t =  F_t(B_t) + \mathbf{\epsilon}_t,
\ee
where $\mathbf{r}_t$ are asset returns, $F_t:\mathbf{R}^{K}\rightarrow \mathbf{R}$ is a differentiable non-linear function that maps the $i^{th}$ row of $B$ to the $i^{th}$ asset return at time $t$. The map is assumed to incorporate a bias term so that $F_t(\mathbf{0})=\alpha_t$. In the special case when $F_t(B_t)$ is linear, the map is $F_t(B_t)=B_t\mathbf{f}_t$. $\mathbf{\epsilon}_t$ is assumed i.i.d., but not necessarily Gaussian.

Equation \ref{eq:non-linear} estimates the conditional mean as $\mathbb{E} [\mathbf{r}_t  |B_t ]= F_t(B_t)$. Hence, we seek to predict returns $\mathbf{r}_t$ at time t with the beta covariates $B_t$, which are by construction measurable at time $t-1$. This model can hence be mapped on to a growing body of research on returns prediction using neural networks. One appealing aspect of this approach is that stationarity of the factor realizations is not required since we only predict one period ahead. 

We approximate a non-linear map, $F_t(B_t)$, with a feedforward neural network cross-sectional factor model:
\be
\mathbf{r}_t =  F_{W_t,b_t}(B_t) + \mathbf{\epsilon}_t,
\ee
where $F_{W_t,b_t}$ is a deep neural network with $L$ layers, that is, a super-position of univariate \emph{semi-affine} functions, $\act^{(\ell)}_{W_t^{(\ell)},b_t^{(\ell)}}$, to give 
\be
\hat{\mathbf{r}}_t:=F_{W_t,b_t}(B_t)=(\act^{(L)}_{W_t^{(L)},b_t^{(L)}}\circ\dots\circ \act^{(1)}_{W_t^{(1)},b_t^{(1)}})(B_t),
\ee
and the unknown parameters are a set of weight matrices $W_t=(W_t^{(1)},\dots, W_t^{(L)})$ and a set of bias vectors $b_t=(b_t^{(1)},\dots, b_t^{(L)})$.  Any weight matrix $W^{(\ell)}\in \mathbf{R}^{m\times n}$, can be expressed as $n$ column m-vectors $W^{(\ell)}=[\mathbf{w}^{(\ell)}_{,1},\dots, \mathbf{w}^{(\ell)}_{,n}]$. We denote each weight as $w^{(\ell)}_{ij}:=\left[W^{(\ell)}\right]_{ij}$.

The $\ell^{th}$ semi-affine function is itself defined as the composition of the activation function, $\act^{(\ell)}(\cdot)$, and an affine map:
\be
\act^{(\ell)}_{W_t^{(\ell)},b_t^{(\ell)}}(Z^{(\ell-1)}):=\act^{(\ell)}\left(W_t^{(\ell)}Z^{(\ell-1)} + b_t^{(\ell)}\right),
\ee
where $Z^{(\ell-1)}$ is the output from the previous layer, $\ell-1$. The activation functions, $\act^{(\ell)}(\cdot)$, e.g. $\act^{(\ell)}(\cdot)=\max(\cdot,0)$, are critical to non-linear behavior of the model. Without them, $F_{W_t,b_t}$ would be a linear map and, as such, would be incapable of capturing interaction effects between the inputs. This is true even if the network has many layers.

\section{Factor Interpretability}\label{sect:interp}
Once the neural network has been trained, a number of important issues surface around how to interpret the model parameters. This aspect is by far the most prominent issue in deciding whether to use neural networks in favor of other machine learning and statistical methods for estimating factor realizations, sometimes even if the latter's predictive accuracy is inferior. In this section, we shall introduce and analyse a method for interpreting multi-layer perceptrons which imposes minimal restrictions on the neural network design.  

\subsection{Sensitivity Approach} We use a gradient-based technique for determining the importance of the input variables.   The method is directly consistent with how coefficients are interpreted in linear regression, i.e. as \emph{model} sensitivities and is thus appealing to asset managers. In our approach, the model sensitivities are the partial derivatives of the fitted model output w.r.t. each input. This method is consistent with \cite{enguerr2019significance} who develop statistical tests for the significance of the factors in a deep network, using their partial derivatives. The test statistic is based on a weighted distribution of the square of the neural network partial derivative, w.r.t. each input. Without a bound on the gradient, everywhere, the integral for their test statistic is unbounded and hence the test statistic could be unbounded.

The scope of their study is limited to a single-hidden layer network and treats the asymptotic distribution of the network as a Gaussian process. Here we shall consider deep networks with an arbitrary number of layers and do not rely on asymptotic approximations, which may be limited when the network has a small number of neurons. Indeed we derive closed form expressions for the Jacobian and Hessian - the off-diagonals provide sensitivities to interaction terms. In contrast to \cite{enguerr2019significance}, our goal is to characterize the dispersion of the empirical sensitivity distribution - confidence intervals of the distributions can be found by Bootstrap sampling (see, for example, \cite{doi:10.1177/1094428105280059}). Such an approach is only useful in practice if the variance of these distributions are bounded. Our primary theoretical result then is to characterize the upper bound on the variance of the partial derivatives and show that it is bounded for finite weights.




To evaluate fitted model sensitivities analytically, we require that the function $\hat{Y}=F(X)$ is continuous and differentiable everywhere. Furthermore, for stability of the interpretation, we shall require that $F(x)$ is a Lipschitz continuous\footnote{If Lipschitz continuity is not imposed, then a small change in one of the input values could result in an undesirable large variation in the derivative}. That is, there is a positive real constant $K$ s.t. $\forall x_1, x_2 \in \mathbb{R}^p$, $|F(x_1) - F(x_2)| \leq K|x_1-x_2|$. Such a constraint is necessary for the first derivative to be bounded and hence amenable to the derivatives, w.r.t. to the inputs, providing interpretability. 

Fortunately, provided that the weights and biases are finite, each semi-affine function is Lipschitz continuous everywhere. For example, the function $\tanh(x)$ is continuously differentiable with derivative, $1-\tanh^2(x)$, is globally bounded.  With finite weights, the composition of $\tanh(x)$ with an affine function is also Lipschitz. Clearly ReLU$(x):=\max(\cdot,0)$ is not continuously differentiable and one can't use the approach described here.  
In a linear regression model
\be
\hat{Y} =F_{\mathbf{\beta}}(X):= \beta_0 + \beta_1X_1 + \dots +\beta_K X_K,
\ee
the model sensitivities are $\partial_{X_i} \hat{Y}=\beta_i$. In a feedforward neural network, we can use the chain rule to obtain the model sensitivities 
\be
J:=\partial_{X} \hat{Y} =W^{(L)}J(I^{(L-1)})=W^{(L)}D(I^{(L-1)})W^{(L-1)}\dots D(I^{(1)})W^{(1)},
\ee
where $D_{ii}(I):=\act'(I_i),~D_{ij}=0,~i\neq j$ is a diagonal matrix.
\subsection{Interaction Effects}  The pairwise interaction effects, frequently used in factor modeling, are readily available by evaluating the elements of the Hessian matrix. For a L layer network, we define the $(i,j)^{(th)}$ element of the Hessian as 
\be
\partial^2_{X_iX_j} \hat{Y}=\sum_{\ell=1}^{L-1} H_{i,j,\ell},~ H_{i,j,\ell}:=W^{(L)}D^{(L-1)}W^{(L-1)}\dots \partial_{X_j} D^{(\ell)}W^{(\ell)}\dots w_i^{(1)}.
\ee
where it is assumed that the activation function is at least twice differentiable everywhere, e.g. $tanh(x)$, softplus etc.

\subsection{Bounds on the Variance of the Jacobian} \label{sect:jac}
Consider a $ReLU$ activated single layer feed-forward network. In matrix form, with $h(x)=\max(x,0)$, the Jacobian, $J$, can be written as a linear combination of Heaviside functions:
\be
J:=J(X)=\partial_{X} \hat{Y} (X)=W^{(2)}J(Z^{(1)})=W^{(2)}H(W^{(1)}X + b^{(1)})W^{(1)},
\ee
where $H_{ii}(Z)=H(Z^{(1)}_i)=\mathbbm{1}_{Z^{(1)}_i>0}$, $H_{ij}=0, ~j\geq i$.

\begin{figure}[h!]
\centering
\includegraphics[width=0.6\textwidth]{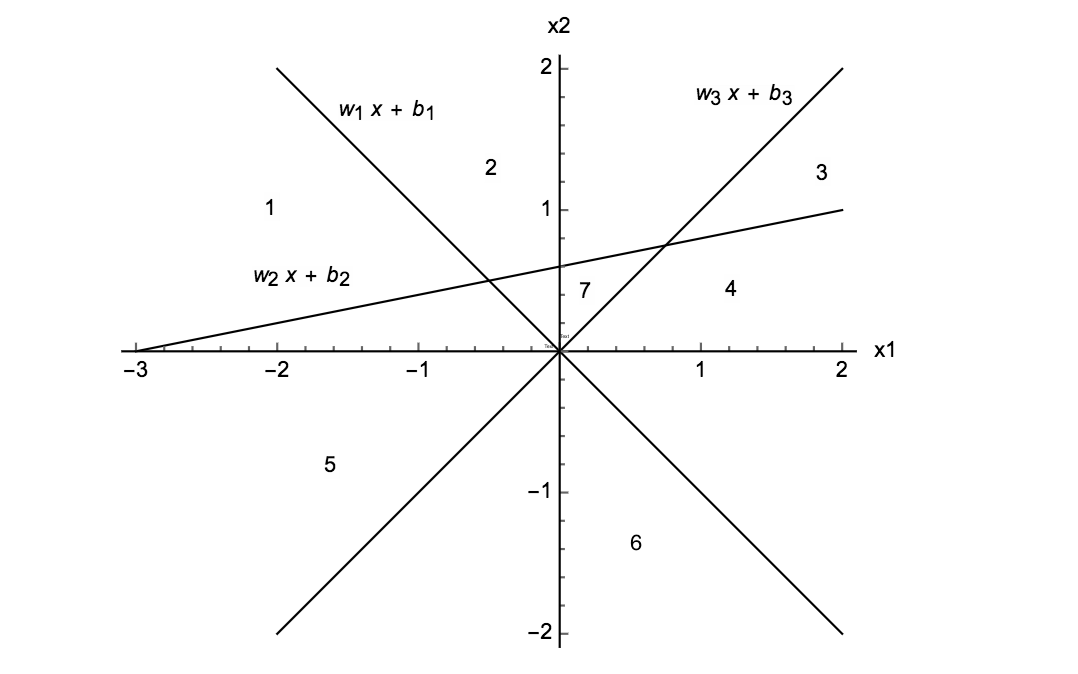}
\caption{\textit{Hyperplanes defined by three neurons in the hidden layer, each with ReLU activation functions, form a hyperplane arrangement. An arrangement of $3$ hyperplanes in $\mathbb{R}^2$ has $n_c\leq \sum_{j=0}^2  {3 \choose j}=7$ convex regions.}}
\label{fig:partition}
\end{figure}

We recall a well-known result, namely the set of $n$ hyperplanes defines a hyperplane arrangement. Such an arrangement of $n\geq p$ hyperplanes in $\mathbb{R}^p$ has $n_c\leq \sum_{j=0}^p {n \choose j}$ convex regions over a bounded domain \citep{2014arXiv1402.1869M}. 
For example in a two dimensional input space, three neurons with ReLU activation functions will divide the
space into $n_c\leq \sum_{j=0}^2  {3 \choose j}=7$ regions, as shown in Figure \ref{fig:partition}. 
We label each of the convex regions in layer $\ell$ as $\{A_1^{(\ell)}, \dots, A_{n_c^{(\ell)}}^{(\ell)}\}$. We can now state the following bound on the Jacobian of the network.
'\begin{thm}[Dixon \& Polson, 2020]
If $X\in \mathbb{R}^p$ is i.i.d. and there are $\{n^{(1)}, \dots, n^{(L-1)}\}$ hidden units across $L-1$ ReLU activated hidden layers, with at most $\{n^{(1)}_c, \dots, n^{(L-1)}_c\}$ convex regions in the bounded hyperplane arrangements of each layer,  then the variance of the Jacobian has the following upper bound:
\be
\mathbb{V}[J_{ij}] \leq \frac{1}{2} ||a_{ij}||^2_2, 
\ee
where $a_{ij}$ is a $(L-1)$ rank tensor with norm $||\cdot||$ whose $(m^{(1)},\dots, m^{(L-1)})$ component is the product over $\ell \in \{1,\dots,L-1\}$ layers of weighted sums of $n^{(\ell)}$ Heaviside functions over convex region $A_{m^{(\ell)}}^{(\ell-1)}$ \& $a_{i,j,m^{(1)},\dots, m^{(L-1)}} \leq 
 w^{(L)}_{i,k^{(L-1)}}w^{(L-1)}_{k^{(L-1)},k^{(L-2)}}\dots w^{(1)}_{k^{(1)},j}$, $\forall m^{(\ell)} \in\{1,\dots, n_c^{(\ell)}\}$. See Appendix for the proof.
\end{thm}

Note that the result holds without (i) distributional assumptions on $X$ other than i.i.d.; and (ii) without specifying the number of data points.

\section{Russell 1000 Factor Modeling}\label{sect:results}
This section presents the application of our framework to the fitting of a fundamental factor model. The Barra factor model includes many more explanatory variables than used in our experiments below, but the purpose, here, is to illustrate the application of our framework to a larger dataset.

For completeness, we provide evidence that our deep neural network factor model generates positive and higher information ratios than linear and quadratic regressions when used to sort portfolios from a larger universe, using 49 factors from Bloomberg - 19 of which are fundamental factors and the remainder are GICS sector dummy variables (see repository documentation for a description of the factors), over an approximately 30 year period of monthly updates between January 1989 to November 2018, and with a coverage universe of 3290 stocks from the Russell 1000 index. All stocks with missing factor exposures are removed the estimation universe for the date of the missing factors only. To avoid excessive turn-over in the estimation universe over each consecutive period, we include all dropped symbols in the index over the 12 consecutive monthly periods. Further details of the data preparation, including sanitization to avoid violating licensing agreements, are provided in the documentation for the source code repository referenced in the abstract.

Training and testing is alternated each period. For example, in the first historical month of the data, the model is fitted to the factor exposures and monthly excess monthly returns over the next period. Cross-validation is performed using this cross-sectional training data, with approximately 1000 symbols. Once the model is fitted and tuned, we then apply the model to the factor exposures in the next period, t+1, to predict the excess monthly returns over [t+1, t+2]. The process is repeated over all periods. Note, for computational reasons, we can avoid cross-validation over every period and instead stride the cross-validation, every other say 10 periods, relying on the optimal hyper-parameters from the last cross-validation periods for all subsequent intermediate periods. In practice we find that performing cross-validation in the first period only is adequate.

Three fold cross-validation over $\{50,100,200\}$ hidden units per layer, $\lambda_1\in\{0,10^{-3},10^{-2}, 10^{-1}\}$ and $\{1,2,3\}$ hidden layers is performed. We find that the optimal architecture has two hidden layers, 100 units per layer, no $L_1$ regularization, and ReLU activation. For an interpretable model, tanh activation functions are required to provide sufficient smoothness (with 50 hidden units per layer being optimal). The linear regression is an OLS model with an intercept. The LASSO model includes quadratic powers in the factors and pairwise interaction terms. The LASSO model uses cross-validation to optimize the $L_1$ regularization parameter and iterative fitting, with 50 alphas, along the regularization path.

To evaluate the relative advantage of the NN in comparison to linear and quadratic regressions, we construct an equally weighted portfolio 
of $n$ stocks with the highest predicted monthly returns in each period. The portfolio is reconstructed each period, always remaining equal but varying in composition. This is repeated over the most recent 10 year period in the data. We then estimate the information ratios from the mean and volatility of the excess monthly portfolio returns, using the Russell 1000 index as the benchmark.

\begin{figure}[h!]
\centering
\begin{tabular}{cc}
\includegraphics[width=0.46\textwidth]{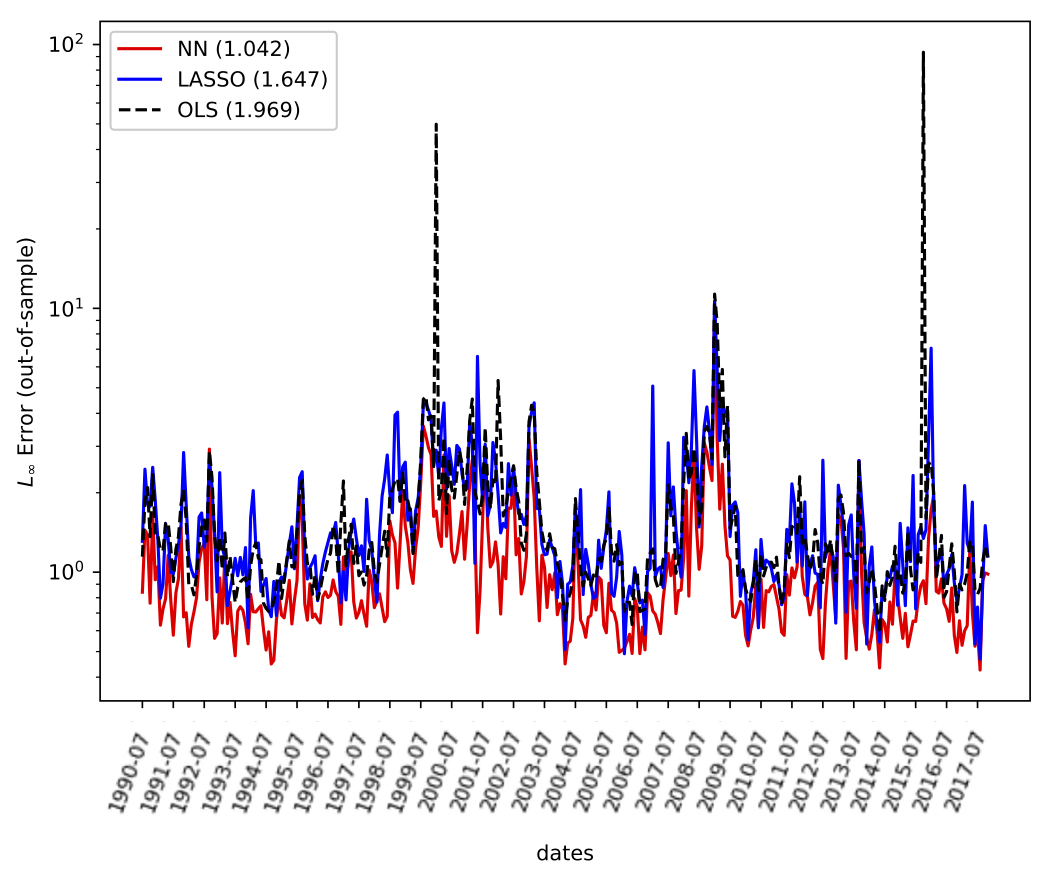} & \includegraphics[width=0.48\textwidth]{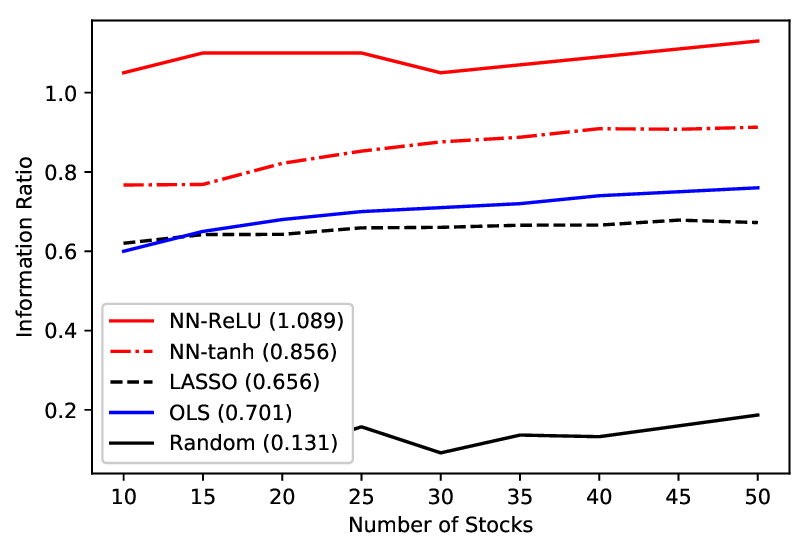}\\
(a) $L_\infty$ error & (b) Information Ratios\\
\end{tabular}
\caption{\textit{(a) The out-of-sample stock returns error, under the $L_\infty$ norm, is compared between OLS, LASSO, and a two-hidden layer deep network applied to a coverage universe of 3290 stocks from the Russell 1000 index over the period from January 1990 to November 2018. The average $L_\infty$ error is shown in parenthesis.  (b) The information ratios of a portfolio selection strategy which selects the $n$ stocks from the universe with the highest predicted monthly returns. The information ratios are evaluated for various portfolios whose number of stocks are shown by the x-axis.}}
\label{fig:errors}
\end{figure}

Figure \ref{fig:errors}(a) compares the out-of-sample performance of neural networks, LASSO, and OLS regression using the $L_\infty$ norm of stock return errors over the coverage universe. We observe the ability of the neural network to capture outliers, with the $L_\infty$ norm of the error in the NN being an order of magnitude smaller than in the OLS model at two dates, 2000-01-01 and 2015-10-01.  The LASSO model exhibits several mid-size outliers which aren't present in the NN model and is outperformed by NN. The average $L_\infty$ norms over all periods is shown in parenthesis and is a factor of 2x smaller for NNs. The $L_\infty$ norm of the OLS error falls to 1.5483 if these two dates are excluded. The out-of-sample MSEs for NNs, LASSO, and OLS are 0.026, 0.029, and 0.254 - the latter decreases to 0.028 with these dates removed. 
Figure \ref{fig:errors}(b) compares the information ratios using the NN, LASSO, and OLS models to identify the stocks with the highest predicted returns. The information ratios are evaluated for equally weighted portfolios with varying numbers of stocks. Also shown, for control, are randomly selected portfolios, without the use of a predictive signal. The mean information ratio for each model, across all portfolios, is shown in parentheses. We observe that the information ratio of the portfolio returns, using the deep learning model (with ReLU), is approximately 1.5x greater than the OLS model. The LASSO model, despite exhibiting lower error than OLS results in a slightly lower information ratio, suggesting that stock selection based on more accurate predicted returns does not directly translate to higher information ratios on account of the stock return covariance. We also observe that the information ratio of the baseline random portfolio is small, but not negligible, suggesting sampling bias and estimation universe modification have a small effect.

\begin{figure}[h!]
\centering
\includegraphics[width=0.97\textwidth]{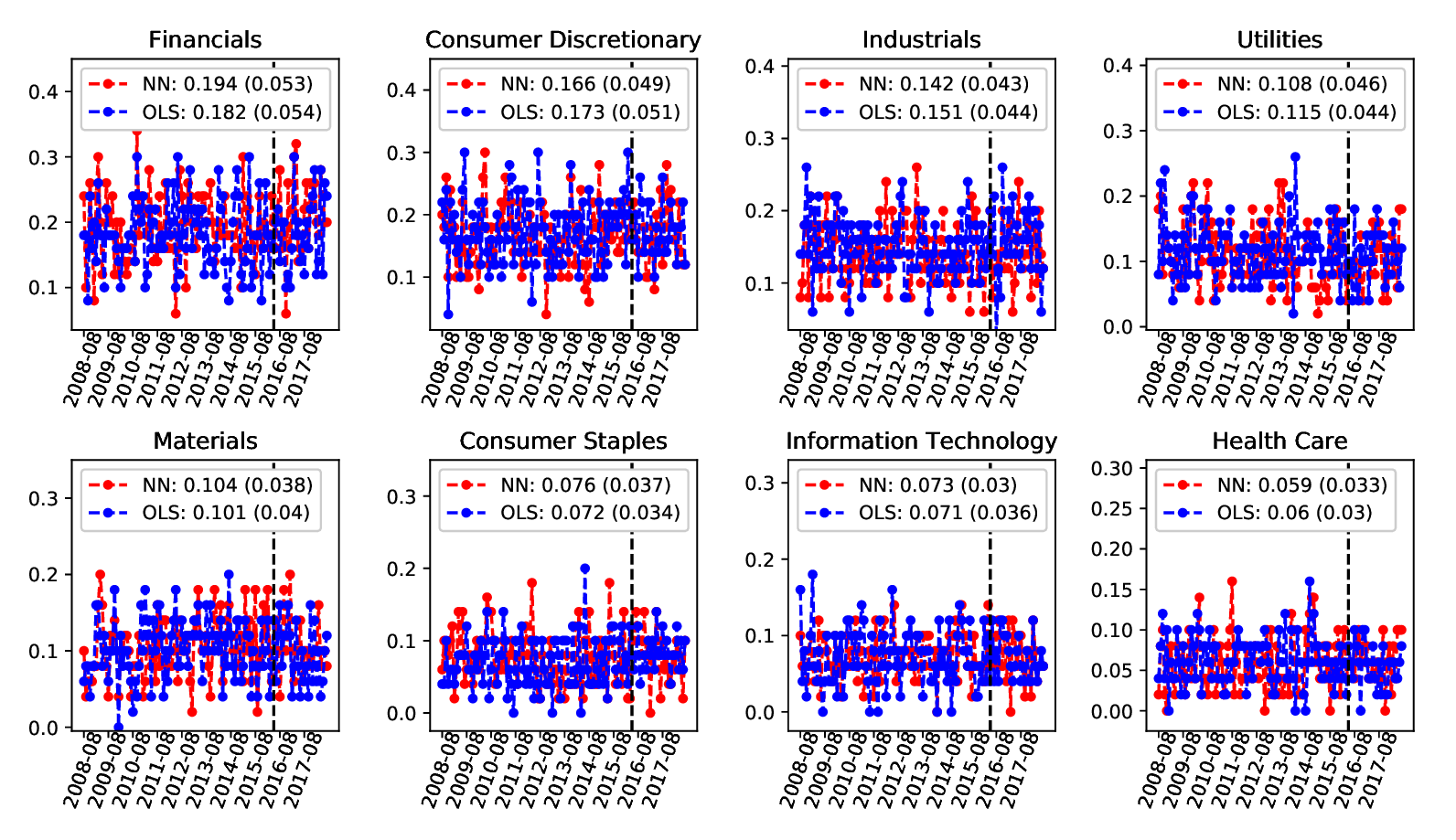}\\
\caption{\textit{The sector tilts are shown over time for each sector, in descending order. The mean and std. devs. of the sector ratios, over the ten year period, are shown in parentheses.}}
\label{fig:sector_tilts}
\end{figure}

Figure \ref{fig:sector_tilts} shows the sector tilts of equally weighted portfolios constructed from the predicted top performing 50 stocks, in each monthly period, over the most recent ten year period in the data. The sectors are ranked by their time averaged ratios, but their tilts vary each month as the portfolios turn-over. 

\begin{figure}[h!]
\centering
\includegraphics[width=0.97\textwidth]{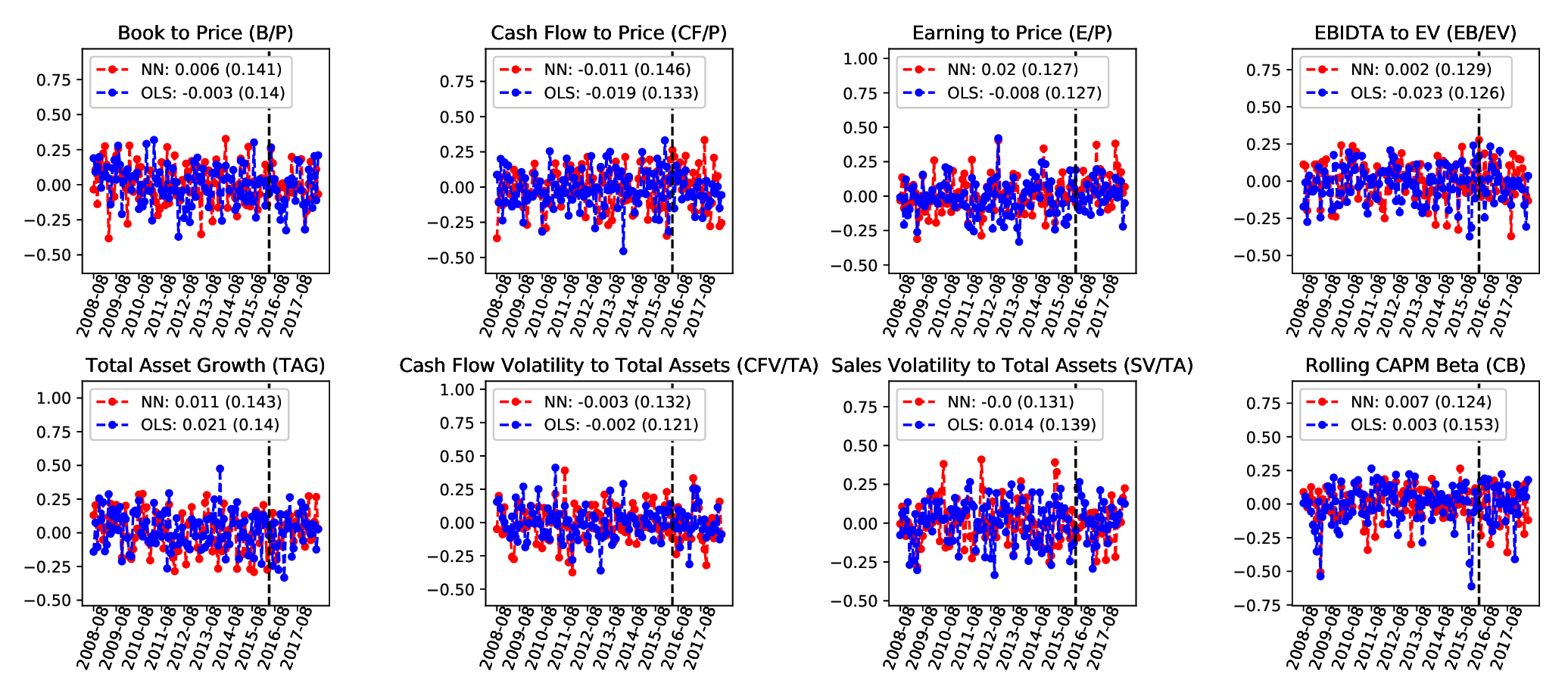}\\
\caption{\textit{The (scaled) factors, averaged over the portfolio, are shown over time for a subset of the factors.}}
\label{fig:factor_tilts}
\end{figure}
Financials is the most dominant sector, with almost 20\% time averaged representation. This is followed by Consumer Discretionary. Note that three of the least representative sectors are excluded: Energy, Communication Services, and Real Estate. The sector tilts across the NN and the OLS are found to be comparable on average. The outlier date, 2015-10-01, is marked with a vertical dashed line-sectors are not useful in explaining the difference. 
Figure \ref{fig:factor_tilts} shows the corresponding (scaled) factors, averaged over the portfolio, for a subset of the factors with non-trivial differences in tilts between OLS and NNs. See Table \ref{tab:r3000} for a description of the factors. In comparison with OLS, we observe that NNs favor assets with higher Book to Price, Earning to Price, and Cash Flow Volatility to Total Assets. OLS favors stocks with higher Total Asset Growth, Sales Volatility to Total Assets, and Rolling CAPM Beta. Note on the outlier date, the NN portfolio overweights Market Cap, Earnings Volatility to Total Assets, Size.

\begin{figure}[h!]
\centering

\includegraphics[width=\textwidth]{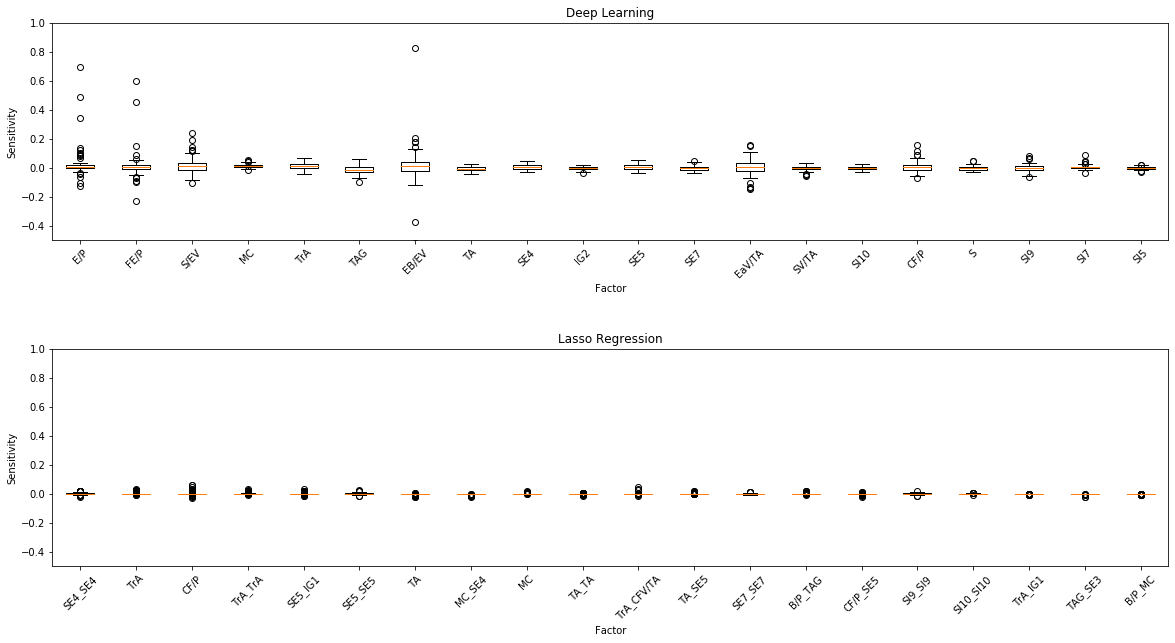}
\caption{\textit{The distribution of factor model sensitivities and interaction terms over the entire ten year period using the deep neural network applied to the Russell 1000 asset factor loadings (top). Earnings/Price (E/P), Forecasted E/P (FE/P) and Sales to Enterprise Value (S/EV) are the three most important factors. The same sensitivities using LASSO are shown (bottom).}}
\label{fig:interaction}
\end{figure}

Figure \ref{fig:interaction} compares the distribution of factor model sensitivities over the entire ten year period using the (top) interpretable NN (i.e. with tanh) and (bottom) LASSO, which includes interaction terms. The sensitivities are sorted in descending order from left to right by their absolute median values and the top 20 are shown. The variance of the NN sensitivity distributions are bounded, a result which is guaranteed by Theorem 3.3.1. 
Terms with $XX\_YY$ denote pairwise interaction effects between factors XX and factors YY. We note that interaction terms feature prominently in LASSO but not in the NN. Trading Activity (TrA), Log Market Cap (MC) and Log (Total Assets) (TA) are the most important factors which are common between the models. We also observe much more variability in the sensitivity of the NN to the most sensitive factors than in LASSO which may explain why the NN is able to capture outliers more effectively. For example, while the interquartile range is moderate, we note that the NN has the capacity to capture outlier sensitivities which are ignored by LASSO.
\section{Summary}\label{sect:summary}
In this paper, we introduce a deep learning framework for fundamental factor modeling which generalizes the linear fundamental factor models. Our framework provides interpretability, with confidence intervals, and ranking of the factor importances and interaction effects. 
In the case when the network contains no hidden layers, our approach recovers a linear fundamental factor model and the framework therefore allows the impact of non-linearity in factors to be assessed. The NN is observed to generate information ratios which are a factor of 1.5x higher than OLS. Furthermore, even if interaction effects are included in a LASSO model, the NN is still observed to better capture outliers on account of its universal representation properties. Future work is needed to investigate the application of the sensitivity based interpretability approach to other types of deep architectures.

\bibliographystyle{chicago}
\bibliography{main}

\appendix
\section{Proof of Theorem}
Recall that a $L$ layer network has the jacobian:

\be
J_{ij}=\sum_{k^{(1)}=1,\dots, k^{(L-1)}=1}^{n^{(1)},\dots, n^{(L-1)}}w^{(L)}_{i,k^{(L-1)}}w^{(L-1)}_{k^{(L-1)},k^{(L-2)}}\dots w^{(1)}_{k^{(1)},j}\prod_{\ell=1}^{L-1} D^{(\ell)}_{k^{(\ell)},k^{(\ell)}}
\ee
which we can write as
\be
J_{ij}=\sum_{k^{(1)}=1,\dots, k^{(L-1)}=1}^{n^{(1)},\dots, n^{(L-1)}}C_{k^{(1)},\dots, k^{(L-1)}}H^{(1),\dots, (L-1)}_{k^{(1)},\dots, k^{(L-1)}}
\ee
where $[C_{k^{(1)},\dots, k^{(L-1)}}]_{ij}:=w^{(L)}_{i,k^{(L-1)}}w^{(L-1)}_{k^{(L-1)},k^{(L-2)}}\dots w^{(1)}_{k^{(1)},j}$ and the product of Heaviside functions is $H^{(1),\dots, (L-1)}_{k^{(1)},\dots, k^{(L-1)}}:=H_{k^{(L-1)}}(I^{(L-1)})\cdot\dots \cdot H_{k^{(1)}}(I^{(1)})$. As before we can rewrite the sum of Heavisides as the sum of indicator functions over convex regions
\be
J_{ij}=\sum_{m^{(1)}=1,\dots, m^{(L-1)}=1}^{n^{(1)}_c,\dots, n^{(L-1)}_c}a_{m^{(1)},\dots, m^{(L-1)}}\prod_{\ell=1}^{L-1} \mathbbm{1}_{Z^{(\ell-1)}\in A^{(\ell)}_{m^{(\ell)}}}, 
\ee
where the sum of Heavisides over the activated convex regions $\{A^{(1)}_{m^{(1)}}, \dots, A^{(L-1)}_{m^{(L-1)}}\}$ is for $Z^{(\ell-1)}\in A^{(\ell)}_{m^{(\ell)}}$:
\be
[a_{m^{(1)},\dots, m^{(L-1)}}]_{ij}:=\sum_{k^{(1)}=1,\dots, k^{(L-1)}=1}^{n^{(1)},\dots, n^{(L-1)}}[C_{k^{(1)},\dots, k^{(L-1)}}]_{ij}\prod_{\ell=1}^{L-1} H^{(\ell)}_{k^{(\ell)}} (Z^{(\ell-1)}).
\ee
The $n^{(\ell)}$ hyperplanes in layer $\ell$ partitions each bounded input space $\mathbb{R}^{n^{(\ell-1)}}$ into at most $n^{(\ell-1)}_c$ convex regions $\{A^{(\ell)}_1, \dots, A^{(\ell)}_{n^{(\ell)}_c}\}$ where $n^{(0)}=p$ and $Z^{(0)}=X$. 
Under expectations we have
\be
\mu_{ij}:=\mathbb{E}[J_{ij}]=\sum_{m^{(1)}=1,\dots, m^{(L-1)}=1}^{n^{(1)}_c,\dots, n^{(L-1)}_c}[a_{m^{(1)},\dots, m^{(L-1)}}]_{ij}p_{m^{(1)},\dots, m^{(L-1)}},
\ee
where the joint probabilities of the event sequence $\{Z^{(0)}\in A^{(1)}_{m^{(1)}},\dots, Z^{(L-2)}\in A^{(L-1)}_{m^{(L-1)}}\}$ implies that
\be
p_{m^{(1)},\dots, m^{(L-1)}}:=\Pr(Z^{(0)}\in A^{(1)}_{m^{(1)}},\dots, Z^{(L-2)}\in A^{(L-1)}_{m^{(L-1)}}),
\ee
if the sequence is assumed unique and an independent set. The variance is given by
\begin{eqnarray*}
\mathbb{V}[J_{ij}]&=&\sum_{m^{(1)}=1,\dots, m^{(L-1)}=1}^{n^{(1)}_c,\dots, n^{(L-1)}_c}[a_{m^{(1)},\dots, m^{(L-1)}}]^2_{ij}p_{m^{(1)},\dots, m^{(L-1)}}(1-p_{m^{(1)},\dots, m^{(L-1)}})\\
&\leq& \frac{1}{2}\sum_{m^{(1)}=1,\dots, m^{(L-1)}=1}^{n^{(1)}_c,\dots, n^{(L-1)}_c}[a_{m^{(1)},\dots, m^{(L-1)}}]^2_{ij}.
\end{eqnarray*}

\clearpage

\section{Description of Factor Model}
\begin{table}[h!]
\caption{\textit{A short description of the factors used in the Russell $1000$ deep learning factor model.}}
\resizebox{\columnwidth}{!}{
\begin{tabular}{l|l|l}
  
\hline
ID& Symbol & \textbf{Value Factors}\\

\hline
1 & B/P & Book to Price\\

2 & CF/P & Cash Flow to Price\\

3& E/P & Earning to Price\\

4 & S/EV & Sales to Enterprise Value (EV). EV is given by \\
 && EV=Market Cap + LT Debt + max(ST Debt-Cash,0), \\
 & & where LT (ST) stands for long (short) term\\

5& EB/EV&   EBIDTA to EV \\

6& FE/P & Forecasted E/P. Forecast Earnings are calculated from Bloomberg earnings consensus estimates data. \\
& & For coverage reasons, Bloomberg uses the 1-year and 2-year forward earnings.\\

17& DIV & Dividend yield. The exposure to this factor is just the most recently announced annual net dividends\\ &&  divided by the market price. \\
&& Stocks with high dividend yields have high exposures to this factor.\\

\hline

& & \textbf{Size Factors}\\

\hline

8 & MC & Log (Market Capitalization)\\

9& S & Log (Sales)\\

10 & TA & Log (Total Assets)\\

\hline

& &  \textbf{Trading Activity Factors}\\

\hline

11& TrA & Trading Activity is a turnover based measure. \\
& & Bloomberg focuses on turnover which is trading volume normalized by shares outstanding. \\
&& This indirectly controls for the Size effect. \\
&&The exponential weighted average (EWMA) of the ratio of shares traded to shares outstanding: \\
&& In addition, to mitigate the impacts of those sharp shortlived spikes in trading volume, \\
&& Bloomberg winsorizes the data: \\
&& first daily trading volume data is compared to the long-term EWMA volume(180 day half-life), \\
&& then the data is capped at 3 standard deviations away from the EWMA average.
\\

\hline

&& \textbf{Earnings Variability Factors}  \\

\hline

12 &EaV/TA & Earnings Volatility to Total Assets. \\
&& Earnings Volatility is measured \\

 && over the last 5 years/Median Total Assets over the last 5 years\\

13 & CFV/TA & Cash Flow Volatility to Total Assets. \\
&& Cash Flow Volatility is measured over the last 5 years/Median Total Assets over the last 5 years\\

14 & SV/TA &  Sales Volatility to Total Assets. \\
&& Sales Volatility over the last 5 years/Median Total Assets over the last 5 year\\

\hline 

& & \textbf{Volatility Factors}\\

\hline

15 & RV& Rolling Volatility which is the return volatility over the latest 252 trading days\\

16 & CB & Rolling CAPM Beta which is the regression coefficient\\
& & from the rolling window regression of stock returns on local index returns\\

\hline

&& \textbf{Growth Factors}\\
\hline
7& TAG & Total Asset Growth is the 5-year average growth in Total Assets\\ && divided by the Average Total Assets over the last 5 years\\
18 & EG & Earnings Growth is the 5-year average growth in Earnings\\ &&  divided by the Average Total Assets over the last 5 years\\
\hline
& & \textbf{GICS sectorial codes}\\
\hline
19-24 & (I)ndustry & $\{10, 20, 30, 40, 50, 60, 70\}$ \\
25-35 & (S)ub-(I)ndustry & $\{10, 15, 20, 25, 30, 35, 40, 45, 50, 60, 70, 80\}$\\
36-45 & (SE)ctor & $\{10, 15, 20, 25, 30, 35, 40, 45, 50, 55, 60\}$\\
46-49 & (I)ndustry (G)roup & $\{10, 20, 30, 40, 50\}$\\
\hline
\end{tabular}
}
\label{tab:r3000}
\end{table}

\end{document}